\pdfoutput=1

\documentclass[11pt]{article}

\usepackage[]{acl}

\usepackage{times}
\usepackage{latexsym}

\usepackage[T1]{fontenc}

\usepackage[utf8]{inputenc}

\usepackage{microtype}

\usepackage{amsmath,amssymb,bm,float,graphicx,multirow,natbib,soul,xspace}
\usepackage{bbm} 
\usepackage{booktabs} 
\usepackage{hyperref} 
\usepackage[mathscr]{eucal}
\usepackage{subcaption} 
\usepackage{wrapfig} 
\usepackage{algorithm,algorithmic}

\title{Detecting Entities in the Astrophysics Literature: A Comparison of Word-based and Span-based Entity Recognition Methods}

\author{Xiang Dai \and Sarvnaz Karimi \\
CSIRO Data61 \\ Sydney, NSW, Australia \\ \{dai.dai;sarvnaz.karimi\}@csiro.au}

\begin{document}
\maketitle
\begin{abstract}
Information Extraction from scientific literature can be challenging due to the highly specialised nature of such text. 
We describe our entity recognition methods developed as part of the DEAL (Detecting Entities in the Astrophysics Literature) shared task. The aim of the task is to build a system that can identify Named Entities in a dataset composed by scholarly articles from astrophysics literature. We planned our participation such that it enables us to conduct an empirical comparison between word-based tagging and span-based classification methods. When evaluated on two hidden test sets provided by the organizer, our best-performing submission achieved $F_1$ scores of $0.8307$ (validation phase) and $0.7990$ (testing phase).
\end{abstract}


\section{Introduction}

A large body of scientific literature is published in different domains, making it difficult for researchers in their respective fields to find information or keep up-to-date. 
Automatic information extraction, in particular Named Entity Recognition (NER), is one of the core methods from the NLP community to assist researchers. It finds mentions of entities of interest in a given text, such as in medicine~\citep{rybinski-csiro-2021-jmir-family-history}, astronomy~\citep{murphy-usyd-2006-alta-astronomy}, geology~\cite{consoli-pucrs-2020-lrec-geoscience}, chemistry~\cite{corbett-boyle-2018-chemlistem}, materials~\citep{friedrich-bosch-2020-acl-sofc-exp} or even finance~\cite{loukas-aueb-2022-acl-finer}. 

Astrophysics scientific literature has its own unique properties, raising some specific challenges for handling of the text. For example, it contains ambiguous names chosen based on the scientists names responsible for a mission or a facility name. While it is not the first time that NER for astrophysics has been studied~\cite{murphy-usyd-2006-alta-astronomy}, it is rather under-studied.
DEAL (Detecting Entities in the Astrophysics Literature) shared task introduced as part of the AACL-IJCNLP 2022 conference has challenged the community with the release of an annotated dataset to pave the way for advancing information extraction methods in this field.

We investigate two different NER methods, word-based tagging and span-based classification, on astrophysics data provided by the organisers of the DEAL shared task. In particular, we examine their effectiveness in extracting 31 different types of entities of interest, such as {\em ComputingFacility} and {\em Wavelength}, and report our experimental results, which led our team to an overall third ranking among 12 teams.


\section{Related Work}

\begin{figure*}[t]
    \centering
    \includegraphics[width=\textwidth]{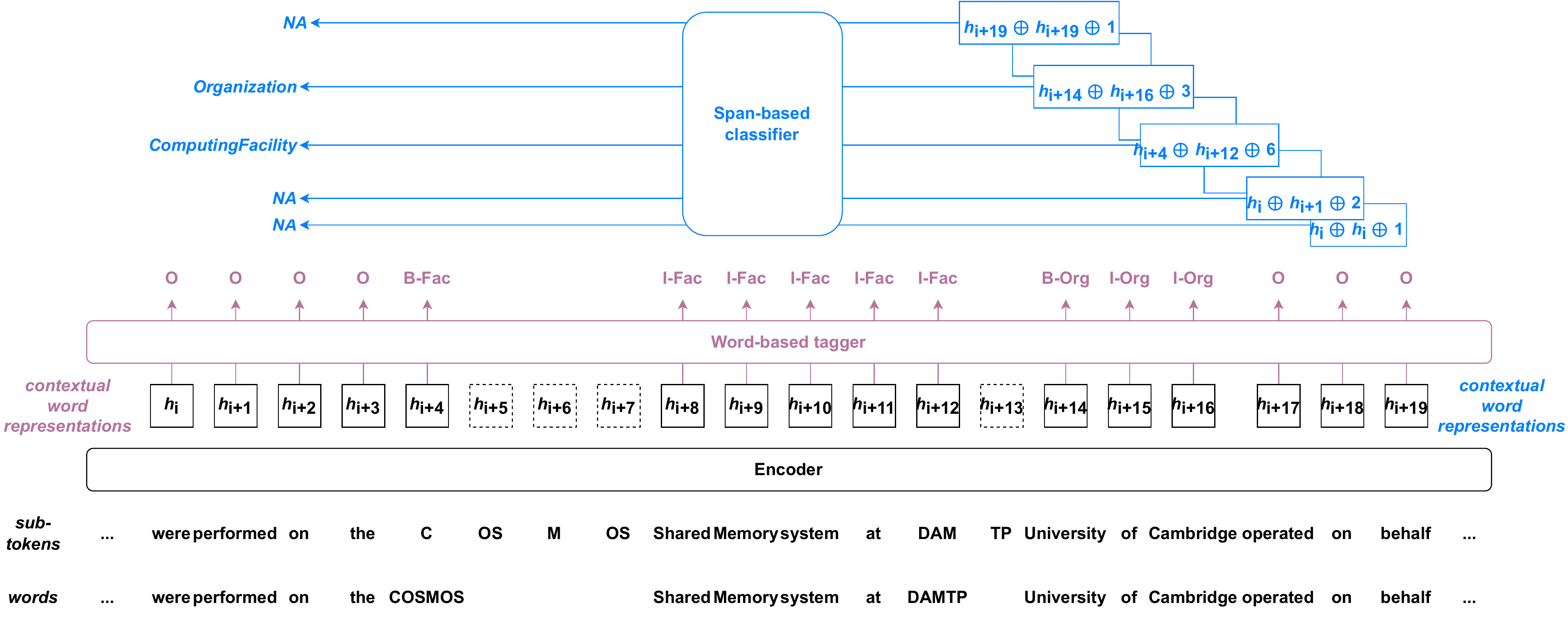}
    \caption{A high-level illustration of word-based and span-based entity recognition methods. These two methods use the same encoder and differ in their classifiers. We use `Fac' to replace `ComputingFacility' for brevity purposes.}
    \label{figure-method}
\end{figure*}

Information extraction, and in particular NER, on scientific literature has attracted substantial research~\citep{augenstein-das-2017-semeval-science-ie,luan-uw-2018-emnlp-scierc,jain-allenai-2020-acl-scirex}. NER refers to the task of identifying mentions of different types of entities in free-text. Types of entities of interest depend on the domain of the text; for example disease names in biomedical text~\cite{dogan-nih-2014-jbi-ncbi-disease,dai-2021-phd-thesis} or numbers in finance~\cite{loukas-aueb-2022-acl-finer}. 
Methods to recognise such entities should also handle different types of the text, including both formal and informal text , such as social media posts~\cite{karimi-csiro-2015-jbi-cadec,basaldella-cambridge-2020-emnlp-cometa}.

For astronomy, there are two existing annotated datasets. 
\citet{hachey-ed-2005-conll-annotation} created {\em The Astronomy Bootstrapping Corpus (ABC)} which is a corpus of 209 annotated article abstracts in English from the radio astronomical papers from the NASA Astrophysics Data System archive. It also includes a further unannotated 778 abstracts used for bootstrapping. 
\citeauthor{hachey-ed-2005-conll-annotation} experimented with active learning for NER on a then novel domain of astronomy.
\citet{murphy-usyd-2006-alta-astronomy} annotated $200,000$ words of text from astronomy articles published on ar{\large X}iv. The dataset is manually annotated with approximately 40 entity types of such as galaxy, star and particle. 
\citeauthor{murphy-usyd-2006-alta-astronomy} also propose a maximum entropy-based NER method on this dataset, reporting an $F_1$ score of approximately $87$\%. 

\citet{grezes-harvard-2021-astrobert} created {\em astroBERT}, a language model for
astronomical text provided by the NASA Astrophysics Data System (ADS).~\footnote{\url{https://ui.adsabs.harvard.edu/}} 
It is pre-trained on $395,499$ English documents from ADS, and is benchmarked for NER, showing improvements over BERT~\cite{devlin-google-2019-naacl-bert}.


\section{Method}

We start from splitting a long document into sentences $\mathcal{S}_1 \mathcal{S}_2 \cdots \mathcal{S}_D$ using a heuristic rule. That is, every full stop is used to mark the end of a sentence if the current sentence consists of more than $10$ words. Given a sentence $\mathcal{S}_i$, two neural entity recognition models are employed to recognize all entity mentions. They use the same encoder (i.e., Transformers~\citep{vaswani-google-2017-neurips-transformer}) and differ in their classifiers, 

A high-level illustration of these two models is shown in Figure~\ref{figure-method}. We describe the encoder in Section~\ref{section-encoder} and two classifiers---word-based tagger and span-based---in Section~\ref{section-token-level-classifier} and Section~\ref{section-span-level-classifier}, respectively.

\subsection{Encoder}
\label{section-encoder}

Sentence words are further split into sub-tokens which can be directly found in the vocabulary~\citep{sennrich-ed-2016-acl-bpe}.
Token embeddings added with position embeddings are taken as input of a stack of Transformer layers~\citep{vaswani-google-2017-neurips-transformer}.
Transformer layer, which consists of self-attention and feed-forward networks, is designed to let tokens interact with each other and thus builds contextual token representations.
In the era of Transformer-based models, model weights (e.g., embeddings, Transformer layers) are usually initialized using publicly available pre-trained models, such as RoBERTa~\citep{liu-fb-2019-roberta}, in this work.

\begin{table*}[tb]
    \centering
    \setlength{\tabcolsep}{3pt}
    \small
    \begin{tabular}{r r c c c c c c c c c c c c}
    \toprule
    & & \multicolumn{4}{c}{Development} & \multicolumn{4}{c}{Validation} & \multicolumn{4}{c}{Testing} \\
    \cmidrule{3-14}
    \bf Method & Encoder & $F_1$ & P & R & MCC & $F_1$ & P & R & MCC & $F_1$ & P & R & MCC \\
     \midrule
    \multirow{4}{*}{Word-based} & \multirow{2}{*}{base} & 0.8158 & 0.8080 & 0.8238 & 0.9124 & 0.8138 & 0.8047 & 0.8230 & 0.9064 & 0.7910 & 0.7958 & 0.7862 & 0.8921 \\
    & & \tiny{(0.0069)} & \tiny{(0.0092)} \vspace{3pt} & \tiny{(0.0053)} & \tiny{(0.0026)} & \tiny{(0.0039)} & \tiny{(0.0059)} & \tiny{(0.0019)} & \tiny{(0.0016)} & \tiny{(0.0038)} & \tiny{(0.0052)} & \tiny{(0.0030)} & \tiny{(0.0018)} \\ 
    & \multirow{2}{*}{large} & 0.8342 & 0.8261 & 0.8424 & 0.9167 & 0.8242 & 0.8191 & \bf 0.8294 & \bf 0.9106 & 0.7985 & 0.8082 & \bf 0.7890 & \bf 0.8959 \\
    & & \tiny{(0.0032)} & \tiny{(0.0006)} & \tiny{(0.0065)} & \tiny{(0.0030)} & \tiny{(0.0048)} & \tiny{(0.0052)} & \tiny{(0.0044)} & \tiny{(0.0013)} & \tiny{(0.0040)} & \tiny{(0.0048)} & \tiny{(0.0034)} & \tiny{(0.0016)} \\
    \midrule
    \multirow{4}{*}{Span-based} & \multirow{2}{*}{base} & 0.8264 & 0.8302 & 0.8227 & 0.9057 & 0.8223 & 0.8326 & 0.8123 & 0.8907 & 0.7996 & \bf 0.8238 & 0.7768 & 0.8760 \\ 
    & & \tiny{(0.0125)} & \tiny{(0.0123)} \vspace{3pt} & \tiny{(0.0130)} & \tiny{(0.0068)} & \tiny{(0.0027)} & \tiny{(0.0013)} & \tiny{(0.0042)} & \tiny{(0.0032)} & \tiny{(0.0004)} & \tiny{(0.0024)} & \tiny{(0.0014)} & \tiny{(0.0015)} \\ 
    & \multirow{2}{*}{large} & \bf 0.8490 & \bf 0.8499 & \bf 0.8482 & \bf 0.9169 & \bf 0.8267 & \bf 0.8328 & 0.8210 & 0.8999 & \bf 0.8034 & 0.8229 & 0.7849 & 0.8837 \\
    & & \tiny{(0.0125)} & \tiny{(0.0050)} & \tiny{(0.0200)} & \tiny{(0.0127)} & \tiny{(0.0019)} & \tiny{(0.0088)} & \tiny{(0.0113)} & \tiny{(0.0042)} & \tiny{(0.0015)} & \tiny{(0.0092)} & \tiny{(0.0101)} & \tiny{(0.0036)} \\
    \midrule
    $1^{st}$& & --- & --- & --- & --- & 0.8364 & 0.8296 & 0.8434 & 0.9129 & 0.8057 & 0.8137 & 0.7979 & 0.8954 \\ 
    $2^{nd}$ && --- & --- & --- & --- & 0.8262 & 0.8145 & 0.8382 & 0.9139 & 0.7993 & 0.8013 & 0.7972 & 0.8978 \\ 
    $3^{rd}$ (ours)&& --- & --- & --- & --- & 0.8307 & 0.8249 & 0.8366 & 0.9138 & 0.7990 & 0.8076 & 0.7906 & 0.8946 \\
    \bottomrule
    \end{tabular}
    \caption{A comparison between word-based and span-based entity recognition models. We report mean scores and standard deviations (in brackets), averaged over three repeats. Shared task results, shown in the bottom, are retrieved from the shared task leaderboard at the end of shared task scoring period. Bold indicates highest number among word- and span-based methods.}
    \label{table-compare-span-and-tagger}
\end{table*}

\subsection{Word-based Tagger}
\label{section-token-level-classifier}

Once we get the contextual representations from the encoder: a list of vectors $h_0, h_1, \cdots, h_n$, where $n$ is the number of sub-tokens in the sentence.
We use the vector corresponding to the first sub-token with each word to represent the word (e.g., $h_{i + 4}$ and $h_{i+12}$ in Figure~\ref{figure-method}).
The word-based tagger takes as input a vector representing one word and outputs a tag which is usually composed of a position indicator and an entity type.
We use BIO position indicators, where B stands for the beginning of a mention, I for the intermediate of a mention, O for outside a mention. For example, {\em COSMOS} in Figure~\ref{figure-method} is assigned a tag {\em B-ComputingFacility}, indicates it is a beginning word of an entity name and its entity type is {e\ ComputingFacility}.

\subsection{Span-based Classifier}
\label{section-span-level-classifier}

We obtain the vector representations for each word in a similar way as described above and then use them to build span representations.
The vectors representing two boundary words and the span length---embedded as a dense vector---are concatenated and taken as input of the span-based classifier.
Note that we use the number of words within the span as span length.
For example, the span length of `COSMOS Shared Memory system at DAMTP' is 6, and the boundary word representations are $h_{i + 4}$ and $h_{i+12}$, shown in Figure~\ref{figure-method}.
The classifier determines whether a span is a valid entity name and what is its entity type.


\section{Dataset and Experimental Setup}
The DEAL shared task organisers released one labelled training set ($1,753$ documents) and one labelled development set ($20$ documents), on which participants can develop their NER systems. Two holdout labelled sets (validation and testing) were used to score submissions, and the labels of these holdout sets were not available to participants until the official scoring period ends. 

The dataset has 31 entity types, with entity `Organization' comprising 16.3\% as highest and entity `TextGarbage', lowest with 0.1\%.
A descriptive statistics of the dataset is shown in Table~\ref{table-data-statistics}.


\begin{table}[t]
    \centering
    \small
    \tabcolsep 3pt
    \begin{tabular}{l r r r r}
    \toprule
     & Training & Development & Validation & Testing \\
     \midrule
\# Documents & 1,753 & 20 & 1,366 & 2,505 \\ 
\# Tokens & 573,132 & 7,454 & 447,366 & 794,739 \\ 
\# Mentions & 41,159 & 628 & 32,916 & 61,623 \\
    \bottomrule
    \end{tabular}
    \caption{The descriptive statistics of the DEAL dataset.}
    \label{table-data-statistics}
\end{table}


We train our models on the first $1,578$ documents of the training set, and the remaining $175$ documents are used for hyper-parameter tuning and best checkpoint selection. We use the Micro-average string match $F_1$ score to evaluate the effectiveness of the models. The model which is most effective on these 175 documents is finally evaluated on the development, validation, and testing sets. 
We repeat all experiments three times using different random seeds, and the mean scores and standard deviations are reported. 

In addition to the $F_1$ score, we report precision (P), recall (R) and Matthew's correlation coefficient (MCC)~\cite{matthews-1975-mcc} metrics, calculated using the scripts provided by the shared task organizers.

\section{Results and Discussion}

We compare word-based and span-based entity recognition models using both RoBERTa-base and RoBERTa-large models.
Results in Table~\ref{table-compare-span-and-tagger} show that span-based model outperforms word-based model by $0.011$ $F_1$ when RoBERTa-base is used, while $0.015$ $F_1$ when RoBERTa-large is used.
From Table~\ref{table-compare-span-and-tagger}, we also observe modest benefit of using RoBERTa-large over RoBERTa-base (0.019 with word-based and 0.023 with span-based).

\paragraph{Task-adaptive pre-training does not guarantee better performance}
Some studies have shown that pre-trained language models are more effective when pre-training data is similar to downstream task data~\citep{dai-csiro-2019-naacl-pretraining-data}.
Task-adaptive pre-training~\citep{howard-ruder-2018-acl-ulmfit,gururangan-allenai-2020-acl-dapt}---continue pre-training on the unlabeled training set for a given task---is a cheap adaptation technique that aims to reduce the disparities between models pre-trained on generic data and domain-specific task data.

\begin{figure}[t]
    \centering
    \includegraphics[width=\linewidth]{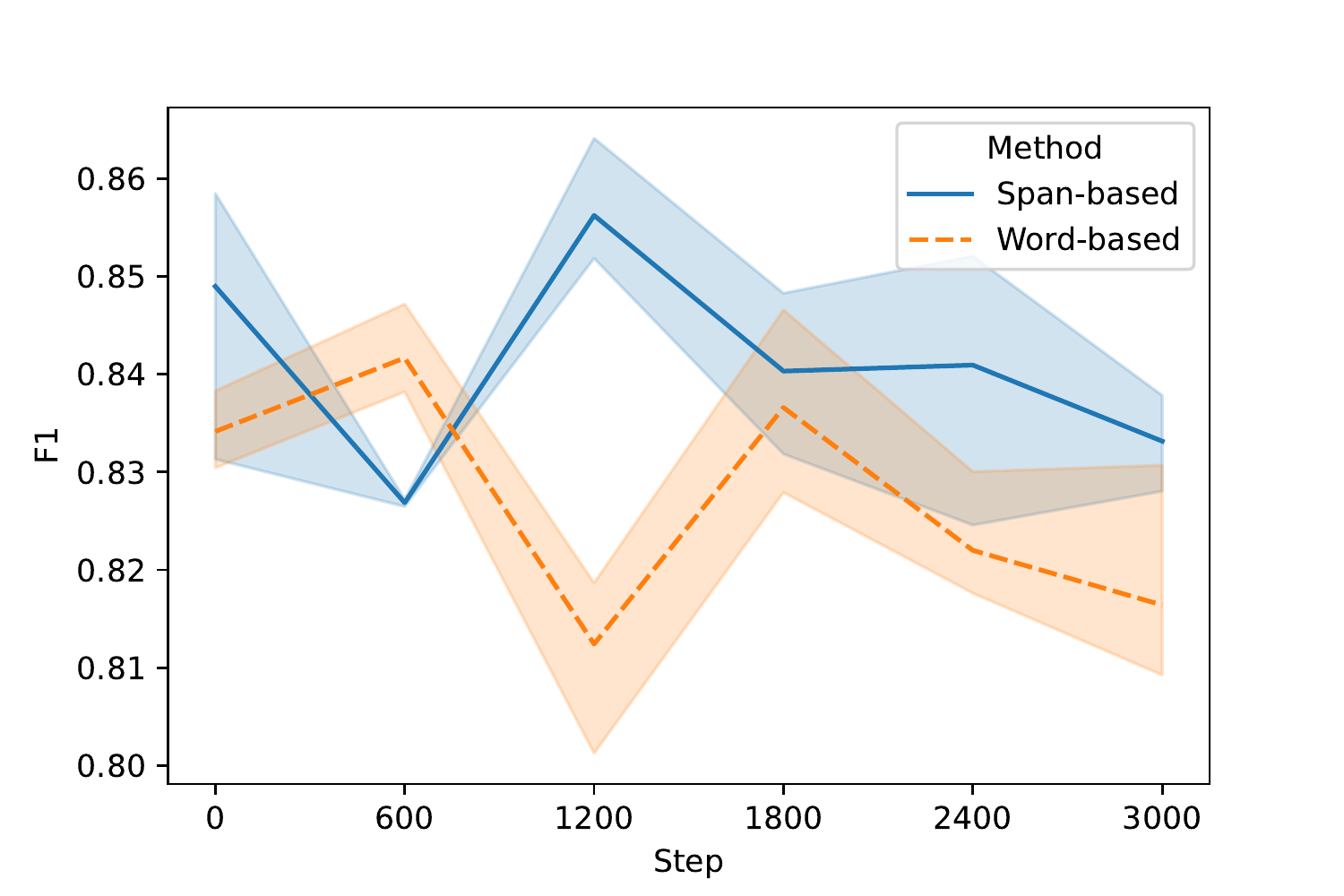}
    \caption{$F_1$ scores evaluated on the development set when task-adaptive pre-trained checkpoints are used. Step $0$ means the vanilla RoBERTa-large is used.}
    \label{figure-tadp}
\end{figure}

We continue pre-training RoBERTa-large on the DEAL training set using masked language modeling.
The total number of optimization steps is 3,000 ($\approx$ 100 epochs), and we save checkpoints every 600 steps.
During the task-adaptive pre-training stage, we observe both the training and development losses keep decreasing, however, the resulting task-adaptive pre-trained checkpoints seem to be very unstable and do not guarantee improved effectiveness (Figure~\ref{figure-tadp}).
Note that \citet{gururangan-allenai-2020-acl-dapt} reported improved effectiveness via task-adaptive pre-training RoBERTa-base, whereas we use RoBERTa-large.
We conjecture the observed instability may be attributed to the optimization difficulties discussed by~\citet{mosbach-saarland-2021-iclr-finetune}, when continue training large size models on small data.


\begin{figure}[b]
    \centering
    \includegraphics[width=\linewidth]{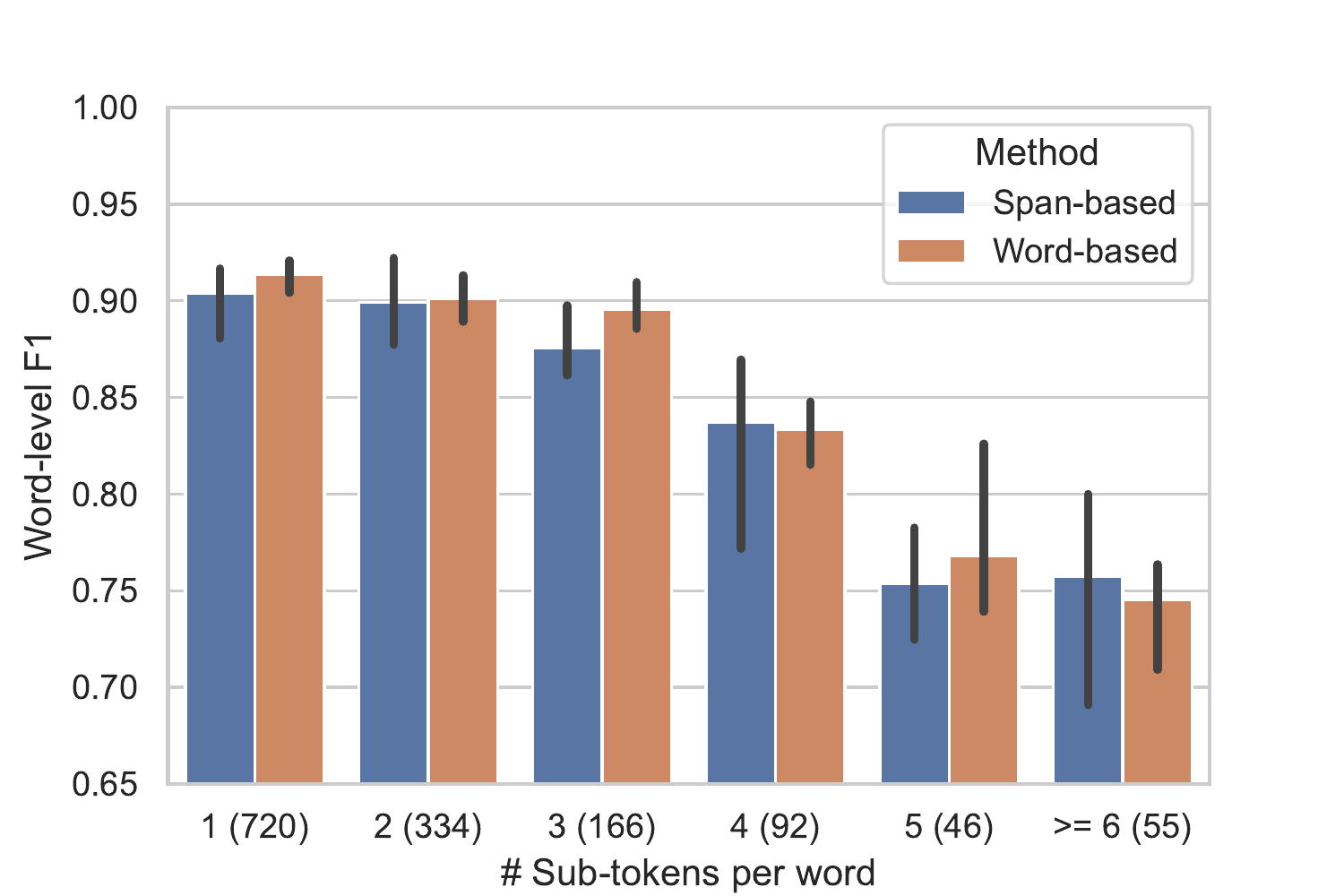}
    \caption{Word-level $F_1$ scores calculated on words that belong to entity names. Number in brackets are the number of corresponding words.}
    \label{figure-over-segmentation}
\end{figure}

\paragraph{Errors due to over-segmentation}
One problem we observe is that many domain-specific terminologies are split into multiple sub-tokens and then taken as input to the encoder.
Taking the sentence in Figure~\ref{figure-method} as an example, since the term `COSMOS' is not in the vocabulary associated with the RoBERTa pre-trained models, it is split into four sub-tokens: `C', `OS', `M', and `OS'.

We calculate the fragmentation ratio---the total number of sub-tokens divided by the total number of words---on the training set of DEAL.
The result, $1.380$, is much higher than the ones calculated on clinical notes ($1.233$) and legal documents ($1.118$) as reported by~\citet{dai-ku-2022-trldc}.
This problem becomes more severe when we only consider words that are part of entity names.
Less than half of these words ($49.8$\%) can be directly found from the RoBERTa vocabulary, and $25.5$\% of words are split into three or more sub-tokens.

We measure the impact of over-segmentation by calculating word-level $F_1$ score on tokens that are part of entity names and grouping words by the number of sub-tokens they are split into.
Figure~\ref{figure-over-segmentation} shows that both word-based and span-based methods suffer from over-segmentation, especially when words are split into three or more sub-tokens.

\begin{table}[t]
    \centering
    \setlength{\tabcolsep}{3pt}
    \begin{tabular}{r c c c}
    \toprule
    & Development & Validation & Testing \\
    \midrule
    Orig & 0.8490 & 0.8267 & 0.8034 \\ 
Innermost & \bf 0.8533 & 0.8293 & 0.8033 \\ 
Outermost & 0.8491 & \bf 0.8298 & \bf 0.8065 \\
\bottomrule
    \end{tabular}
    \caption{The results of applying simple post-processing on outputs from span-based methods. We post-process outputs from span-based model using RoBERTa-large. Bold indicates highest number in the column.}
    \label{table-span-post-processing}
\end{table}

\paragraph{Errors due to nested predictions}
Span-based methods were originally designed to tackle nested NER~\citep{byrne-2007-icsc-nested-ner,ringland-usyd-2019-acl-nne,wang-zju-2020-acl-pyramid}, where two entity names may nest each other.
For example, the span-based method may predict both `COSMOS Shared Memory system' and `COSMOS Shared Memory system at DAMTP' as {\em ComputingFacility} entities.
However, the annotations of DEAL shared task do not allow nested structure.
We find that span-based method benefit from post-processing via resolving these nested predictions.
Results in Table~\ref{table-span-post-processing} show that simple post-processing---keeping only entity names that are not contained by any other names (Innermost) or only entity mentions that do not contain any other names (Outermost)---can bring moderate improvements.

\section{Conclusions}
We reported our experiments on extracting mentions of 31 different types of entities from astrophysics scientific literature. Two different sets of methods based on words and spans were compared. Results show that span-based method using RoBERTa-large pre-trained models outperforms the widely used word-based sequence tagging method.

Potential research directions include building better span representations with the help of external knowledge base; enhancing pre-trained models with domain-specific vocabulary; and, combing the strengths of word-based and span-based models.


\paragraph*{Acknowledgements} This work is supported by The Commonwealth Scientific and Industrial Research Organisation (CSIRO) Precision Health Future Science Platform (FSP). Experiments were undertaken with the assistance of resources and services from the National Computational Infrastructure (NCI), which is supported by the Australian Government.

\bibliography{references}

\end{document}